\documentclass[10pt,twocolumn,letterpaper]{article}

\usepackage{cvpr}              

\usepackage{graphicx}
\usepackage{amsmath}
\usepackage{amssymb}
\usepackage{booktabs}

\usepackage[pagebackref,breaklinks,colorlinks]{hyperref}

\usepackage[capitalize]{cleveref}
\crefname{section}{Sec.}{Secs.}
\Crefname{section}{Section}{Sections}
\Crefname{table}{Table}{Tables}
\crefname{table}{Tab.}{Tabs.}

\def\cvprPaperID{*****} 
\def\confName{CVPR}
\def\confYear{2022}

\begin{document}

\title{Investigating the Robustness and Properties of Detection Transformers (DETR) Toward Difficult Images}

\author{
Zhao Ning Zou$^*$\\
The Australian National University\\
Canberra, ACT\\
{\tt\small ZhaoNing.Zou@anu.edu.au}
\and
Yuhang Zhang$^*$\\
The Australian National University\\
Canberra, ACT\\
{\tt\small Yuhang.Zhang@anu.edu.au}
\and
Robert Wijaya$^*$\\
The Australian National University\\
Canberra, ACT\\
{\tt\small Robert.Wijaya@anu.edu.au}
}
\maketitle
\def\thefootnote{*}\footnotetext{equal contribution}\def\thefootnote{\arabic{footnote}}

\begin{abstract}
Transformer-based object detectors (DETR) have shown significant performance across machine vision tasks, ultimately in object detection. This detector is based on a self-attention mechanism along with the transformer encoder-decoder architecture to capture the global context in the image. The critical issue to be addressed is how this model architecture can handle different image nuisances, such as occlusion and adversarial perturbations. We studied this issue by measuring the performance of DETR with different experiments and benchmarking the network with convolutional neural network (CNN) based detectors like YOLO and Faster-RCNN. We found that DETR performs well when it comes to resistance to interference from information loss in occlusion images. Despite that, we found that the adversarial stickers put on the image require the network to produce a new unnecessary set of keys, queries, and values, which in most cases, results in a misdirection of the network. DETR also performed poorer than YOLOv5 in the image corruption benchmark. Furthermore, we found that DETR depends heavily on the main query when making a prediction, which leads to imbalanced contributions between queries since the main query receives most of the gradient flow.
\end{abstract}

\section{Introduction}
\label{sec:intro}
With the continuous development of deep learning, computer vision has reached a new stage, and target detection, as one of the very vital core directions, has also received attention and many applications based on object detection algorithms.

Before the concept of deep learning was introduced, object detection was mostly based on manual feature extraction. However, as manual feature extraction methods often failed to meet various features in the targets, traditional target detection algorithms could not meet people's needs. After the rise of deep learning, neural networks can automatically learn powerful feature extraction and fitting capabilities from large amounts of data. Thus many DL-based object detectors with excellent performance have emerged. These detectors can be broadly classified into three categories: two-stage object detection, one-stage object detection, and transformer-based object detection.

Faster R-CNN \cite{rcnn} is the most popular two-stage detector nowadays. It first generates a proposal for the object bounding box in the image through a network. It then extracts features from each candidate box, and uses them for object classification and bounding box regression tasks to obtain the final bounding box. On the other hand, the YOLO \cite{yolo} series model is well-known as a one-stage detector, which discards the anchor frame setting in two-stage and extracts the prediction frames directly from the image. Furthermore, with the increasing popularity of transformers applied in computer vision tasks, new transformer-based object detectors, such as DETR \cite{detr}, have also emerged. Instead of using anchor frames and NMS, DETR uses an encoder-decoder structure to classify each object in the image.

DETR is an end-to-end target detection network proposed by Facebook in 2020. Compared to traditional RNNs, DETR uses multiple self-attentive structures, and the parallel computing used therein allows DETR to extract relevance efficiently in context. It is also one of the best-performing target detection methods available.

This paper investigated the properties of DETR and compared the robustness of the network using different interference to the image data, such as sticker and occlusion. We find three main findings in our experiment (1) DETR can handle a small amount of occlusion well compared to Faster R-CNN and YOLOv5, but when there is too much information loss, the attention mechanism of DETR is difficult to be useful. (2) the adversarial patch manages to produce a new set of unnecessary keys, queries, and values in the network, which in most cases, results in the misdirection of the network. (3) The image corruption benchmark performance of DETR is lower than that for YOLO model. (4) We observed a main query phenomenon in the DETR model and showed that is caused the slow convergence problem of the model.


\section{Related Works}
\label{sec:formatting}
Recent successes of Transformer-based models in computer vision tasks have inspired several works \cite{rw1, rw2, rw4, rw5, rw6, c, rw8, rw9, rw10} that study their robustness against corrupted images and adversarial attacks. Some works \cite{rw1, rw2, c, rw9} claimed that transformers are more robust than CNNs in different evaluation settings, including adversarial attacks, while others \cite{rw3, rw6, rw8, rw10} hypothesized that transformers are vulnerable as CNNs. In this study, we aim to understand the robustness of the transformers-based mechanism for object detector settings. We evaluated their robustness against patch masking, adversarial attacks, and common natural corruption images.

Occlusion is one of the effects of target detector performance \cite{e}. There are two general types of occlusion: intra-class occlusion and inter-class occlusion. Many problems such as pedestrian detection, stereo images, etc. face problems caused by occlusion, whether for outdoor or indoor scene tasks. Because when an object is occluded, some information is lost and the remaining information may be difficult to recognise, studying the occlusion problem is an inevitable step for target detection algorithms. \cite{c} evaluated the performance of Vision Transformer on occlusion. Therefore, we also analyzed object detectors performance on occluded images.

Regarding a more realistic robustness benchmark, \cite{bm_1} proposed the ImageNet-C dataset, which consists of 15 diverse corruption types, covering noise, blur, weather, and digital categories. This dataset was first proposed for image classification tasks. In \cite{bm_2}, Claudio M. \emph{et al.} evaluated object detectors' performance in bad weather. They proposed several robustness benchmarks for object detection named Pascal-C, Coco-C, and Cityscapes-C. In our work, we utilized their image corruption generator and focused on the reasons behind the performance drop of object detectors.


\section{Methodology}

\subsection{Occlusion}
Object occlusion is a major problem in target detection applications. Whether it is pedestrian detection, object tracking, or autonomous driving, the object to be detected may be occluded within or between classes. Such partial occlusion reduces the features extracted from the objects and affects the object detector's precision.

Object occlusion has different implications for different tasks. In this paper, we set the value of the image patches that need to be occluded to 0 to reach the information loss effect in that part. Moreover, DETR is a network based on a self-attentive model, so it is essential to study the impact of regions containing more information on DETR. In this paper, we use two different occlusion methods to test Faster R-CNN, YOLO, and DETR, (a) random occluded, (b) salient occluded.

(a)	Random occulted: We use dataset COCO128 to test the anti-jamming capability of DETR. Since the image size is different in COCO128, we split the image into 10 x 10 patches and randomly occlude these patches by setting the pixel value of the occluded patches to 0 to simulate the loss of information. The occlusion ratio is set to the number of occluded patches/total patches. In our experiments, we will test images with masking ratios of 0.2, 0.4, 0.6, and 0.8, respectively.

(b)	Salient occluded: The self-attentive mechanism in the transformer contributes to the excellent performance of DETR. Therefore, it can selectively extract information from the feature map for later classification and box prediction. In a realistic study of how resistant DETR is to interference, the focus should be on the effect of salient regions. In this experiment, we treated the target with different occlusion rates to find the reason why significant regions affect robustness. We considered the area containing the top 20\% of prospective information to be significant. We use different occlusion rates to process the salient part of patches in the object and feed the processed images into the three networks for comparison. Then we use different occlusion rates to process the salient part of patches in the object and feed the processed images into the three networks for comparison.

\subsection{Adversarial Stickers}
The adversarial attack is a popular method to disturb the output precision of machine learning models. By adding a small portion of adversarial perturbation to the image, the detectors can make a misleading judgment, resulting in a significant decrease in performance. Some research utilized a similar approach to evaluate the networks. In \cite{a}, Sharif \emph{et al.} demonstrated that using adversarial glasses is possible to fool facial recognition systems. These glasses were designed to fit any face, allowing it to impersonate any person. Another work \cite{b} demonstrated different methods for constructing fake stop signs to misclassify the models by creating a poster that looks like a stop sign or modifying the stop signs to make it hard to recognizable. This indicated that an adversarial patch is a reliable way to evaluate the performance and robustness of a particular network.

We utilized the attack by completely replacing a portion of the image with a sticker (patch). The stickers are masked with some constraints to fit in different sizes of images by applying random values of translation and scaling to place the stickers in different locations with different sizes. In particular, given a three-dimensional image $x$, patch $p$, location $l$, and transformation $t$ (in this case translation and scaling), we defined a patch operator $S(p,x,l,t)$ which first apply the transformation to the patch, and then following the transformation of patch $p$ to the location $l$ in the image $x$.

This attack proposed to distract the way object detectors
make their predictions and draw the bounding boxes. In the
case where an image contain several objects, the network must be able
to decide the most salient object in the image. Thus, when a targeted object which some of its part overlapped by a sticker, the detectors should be able to classified the object without significantly reducing the accuracy of the targeted object and other objects in the image.

DETR uses transformers to capture the global context of the image. Since the adversarial stickers attack only change the pixel values of a certain region in the image, it is essential to see whether the global scope attention of DETR. Moreover, we want to investigate whether DETR is more robust to this attack than CNN-based mechanism detectors such as YOLO and Faster-RCNN.

\subsection{Benchmark}
\begin{figure}[h]
\par
\raisebox{-.5\height}{\includegraphics[width=8.3cm]{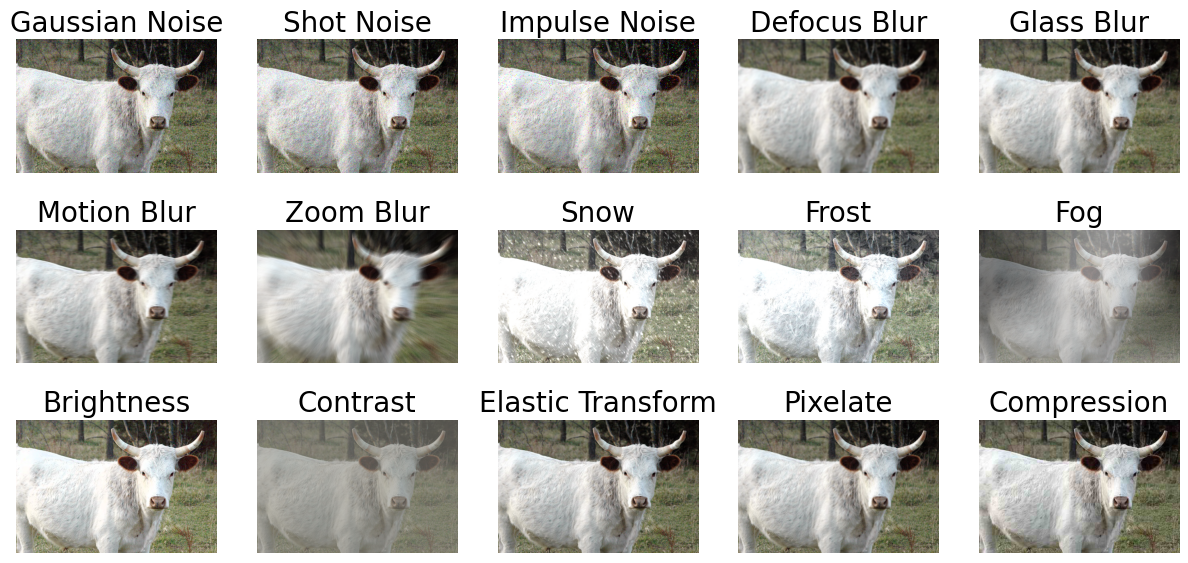}}%
\par
\caption{The image corruptions used in Benchmark evaluation.}
\label{fig:bm_illustration}
\end{figure}

In a real-world scenario, the camera's photo sometimes suffers from common corruptions such as defocus blur, motion blur, or bad weather. Therefore, we wanted to evaluate the performance of YOLOv5m, DETR R50, and DETR R101 on corrupted images. The evaluations were done using 15 image corruptions with five levels of severity in each corruption category, as shown in Figure \ref{fig:bm_illustration}.

Apart from model comparison, we also investigated the reasons behind the performance drop of DETR models. We analyzed the attention maps from the transformer encoder and decoder to see how they changed as the image corruption became more severe.

\subsection{Query Properties}
One of the critical innovations in DETR is the introduction of object queries in the transformer decoder. It is implemented as learnable positional encodings that are added to the decoder input at each layer. These queries help the model decode a set of box coordinates and their class labels. The number of the object query is usually smaller than 100 to reduce the number of bounding boxes generated at the early stage and the computation cost. In \cite{detr}, queries have shown unique preferences on objects with a particular size and position. Therefore it is believed that each query has several modes of operation focusing on different areas and box sizes. In particular, the transformer architectures allow the object query to utilize the context in the whole image to detect large objects more precisely. In this paper, we further analyzed the properties of queries and their relations with class labels. We also presented the observation of main queries and their significant contributions to the DETR's model. Finally, we fine-tuned a DETR R50 model on pascal-voc dataset to investigate the model's transfer learning ability. 


\section{Evaluation}
We record the experiment results with the primary COCO challenge metric, the mean Average Precision (mAP). This metric averages precision-recall scores at different Intersection-overUnion (IoU) threshold. Moreover, to mitigate bias toward local patch, we ignore predicted boxes with less than 50\% with the target box. We did the same for original images for a fair comparison between clean and attacked images.

\subsection{Evaluation on Occlusion}
\begin{figure}[h]
\par
\raisebox{-.5\height}{\includegraphics[width=8.3cm]{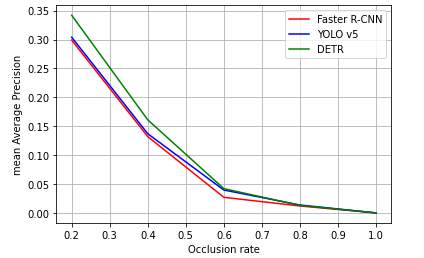}}%
\par
\caption{The mAP of different occlusion rate on DETR, YOLO and Faster R-CNN}
\label{fig:mAP}
\end{figure}

We tested the robustness of Faster R-CNN, YOLOv5, and DETR on the COCO128 dataset with occlusion ratios (number of patches masked/total patches) of 0.2, 0.4, 0.6, and 0.8, and the results are shown in Figure \ref{fig:mAP}. To prevent random perturbations in random occlusion, four separate experiments were conducted, and the evaluation values of the experimental results were taken. DETR outperformed Faster R-CNN and YOLOv5 in several experiments, and the average decrease in accuracy of DETR during the process of increasing the occlusion ratio from 0.2 to 0.8 was $[52.92\%, 73.91\%, 69.04\%]$, while the figures for Faster R-CNN and YOLOv5 were $[55.85\%, 79.54\%, 55.56\%]$ and $[54.93\%, 71.09\%, 65.40\%]$. It can be seen that DETR performs well in terms of resistance to interference from information loss.

\begin{table}[h!]
  \begin{center}
    \label{tab:table1}
    \begin{tabular}{l|c|c|c|c}
      \textbf{Occlusion rate} & 
      \textbf{0.2} & 
      \textbf{0.4} &
      \textbf{0.6} &
      \textbf{0.8} \\
      \hline
      DETR & 0.342 & 0.161 & 0.042 & 0.013\\ 
      YOLOv5 & 0.304 & 0.137 & 0.0396 & 0.0137\\ 
      Faster-RCNN & 0.299 & 0.132 & 0.027 & 0.012\\
    \end{tabular}
  \end{center}
  \caption{The Mean Average Precision (mAP) result comparison of DETR, YOLOv5 and Faster-RCNN on images with different occlusion ratio.}
\end{table}

In the experiments with significant areas occluded, we also treated the significant areas of the images with occluded rates of $0.2, 0.4, 0.6$, and $0.8$, respectively.

\begin{figure}[h]
\par
\raisebox{-.5\height}{\includegraphics[width=8.3cm]{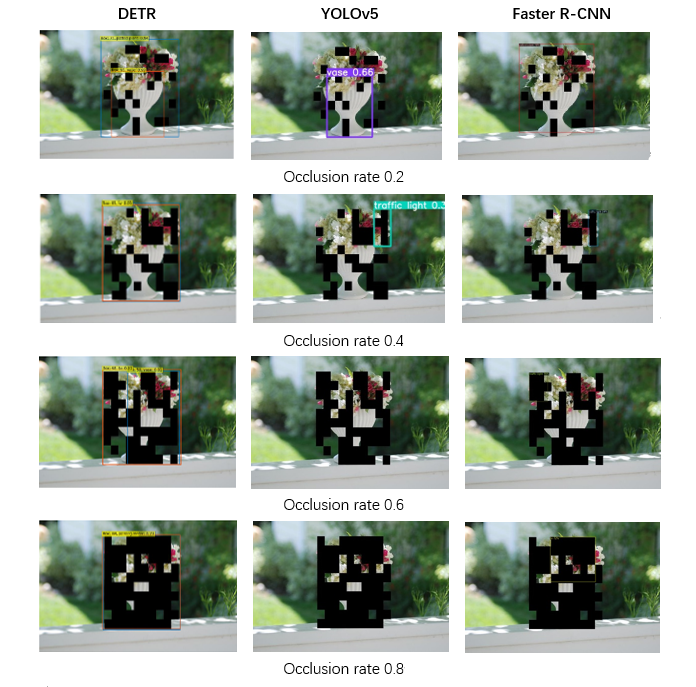}}%
\par
\caption{The detection results of different occlusion rate on DETR, YOLO and Faster R-CNN}
\label{fig:detection results}
\end{figure}

At a significant region occluded rate of 0.2, DETR performed the best, followed by YOLOv5 and Faster R-CNN. For example, DETR detected two classes, potted plant and vase, and both had an accuracy of over 90\%. The other two networks detected only one class of targets with accuracies below 80\%. When the occlusion rate of the salient regions was raised to 0.4 and above, all three networks showed false detections and failed to detect targets. From the results, we can see that DETR can still perform the task of target detection well when a small portion of important information is missing. We illustrate why DETR has better robustness against occlusion by analyzing DETR's attention map.

As can be seen in Figure \ref{fig:detection results}, when the detected object is partially occluded, DETR's attention will mainly focus on the not occluded part because the transformer can combine global information to fill in the loss part by surrounding pixels. However, it can be seen that when too much information is missing due to too many occluded parts, the DETR's attention gradually starts to diverge, and it is unable to find the detected object accurately.

\begin{figure}[h]
\par
\raisebox{-.5\height}{\includegraphics[width=4.1cm]{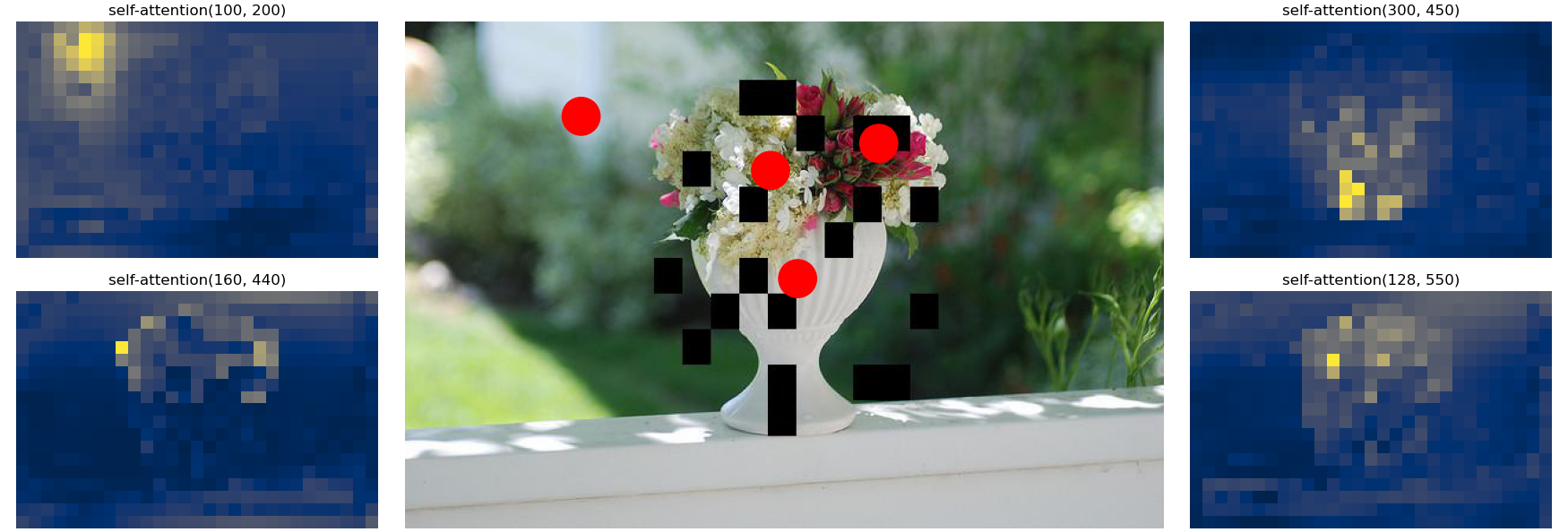}}%
\hfill
\raisebox{-.5\height}{\includegraphics[width=4.1cm]{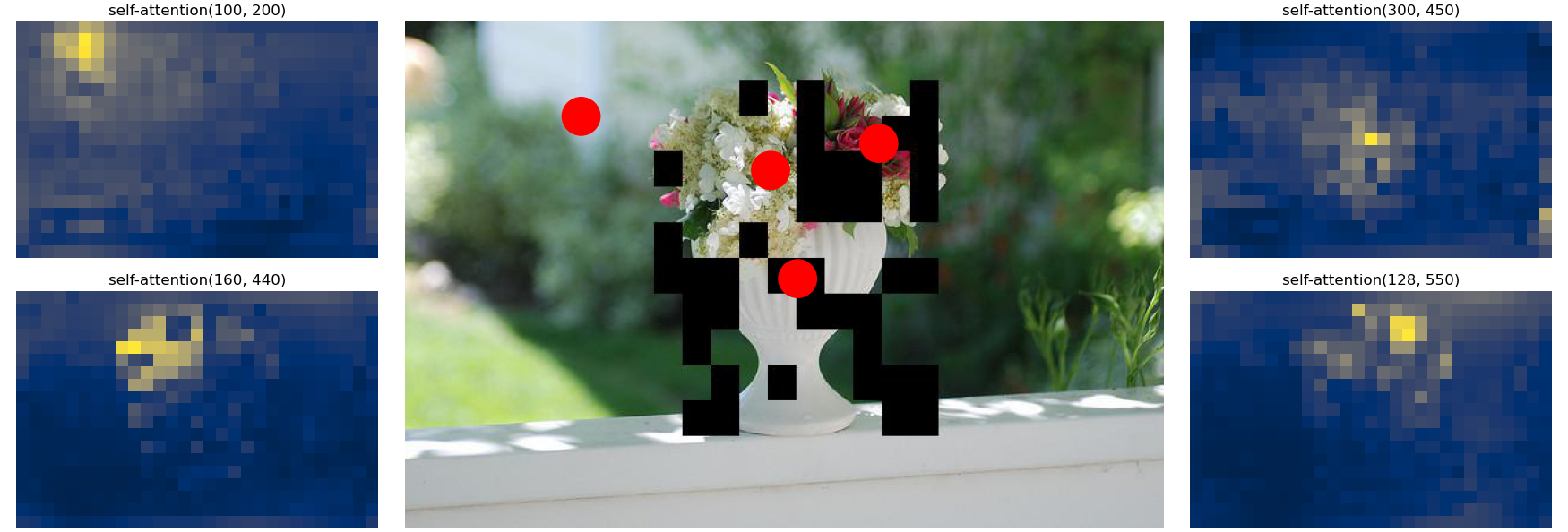}}%
\hfill
\raisebox{-.5\height}{\includegraphics[width=4.1cm]{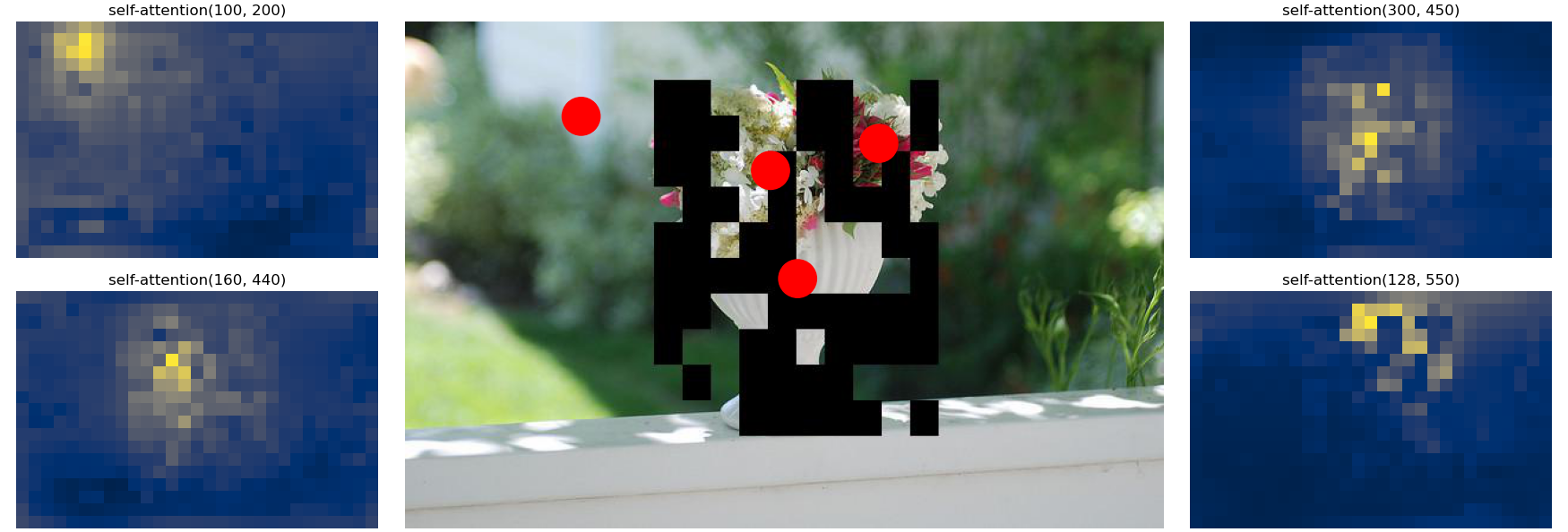}}%
\hfill
\raisebox{-.5\height}{\includegraphics[width=4.1cm]{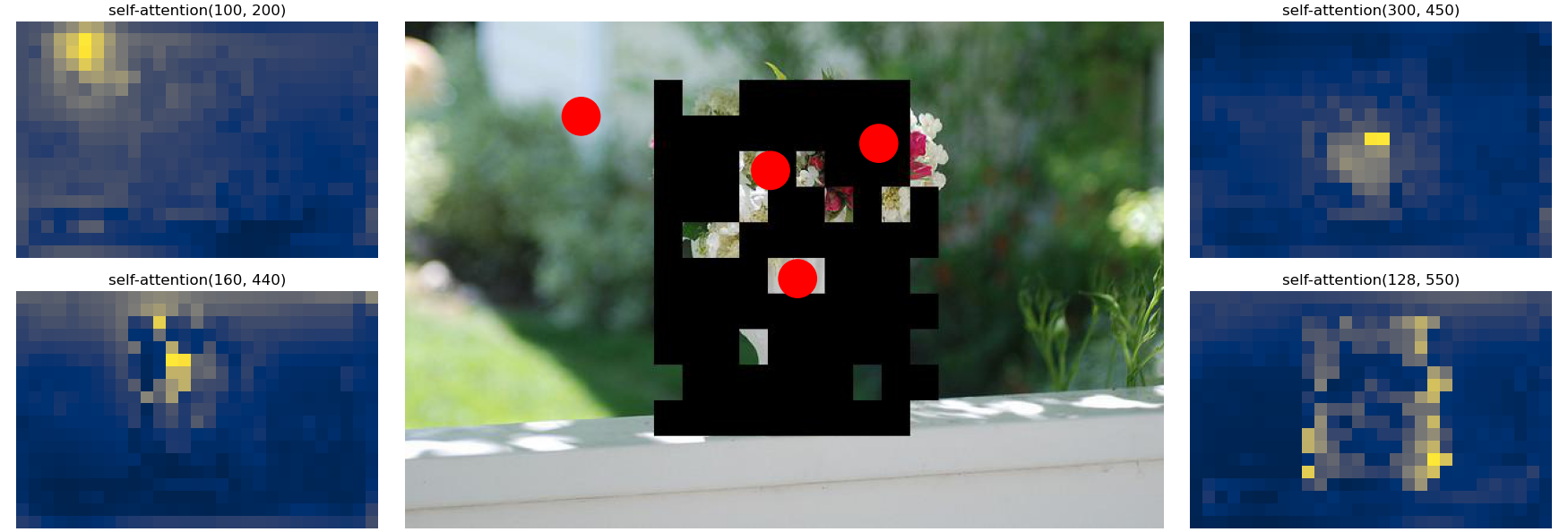}}%
\par
\caption{DETR detection results on different occlusion rate 0.2 (left top), 0.4 (right top), 0.6 (left bottom), 0.8 (right bottom).}
\label{fig:detection results}
\end{figure}

\subsection{Evaluation on Adversarial Stickers}
\begin{figure}[h]
\par
\raisebox{-.5\height}{\includegraphics[width=4.1cm]{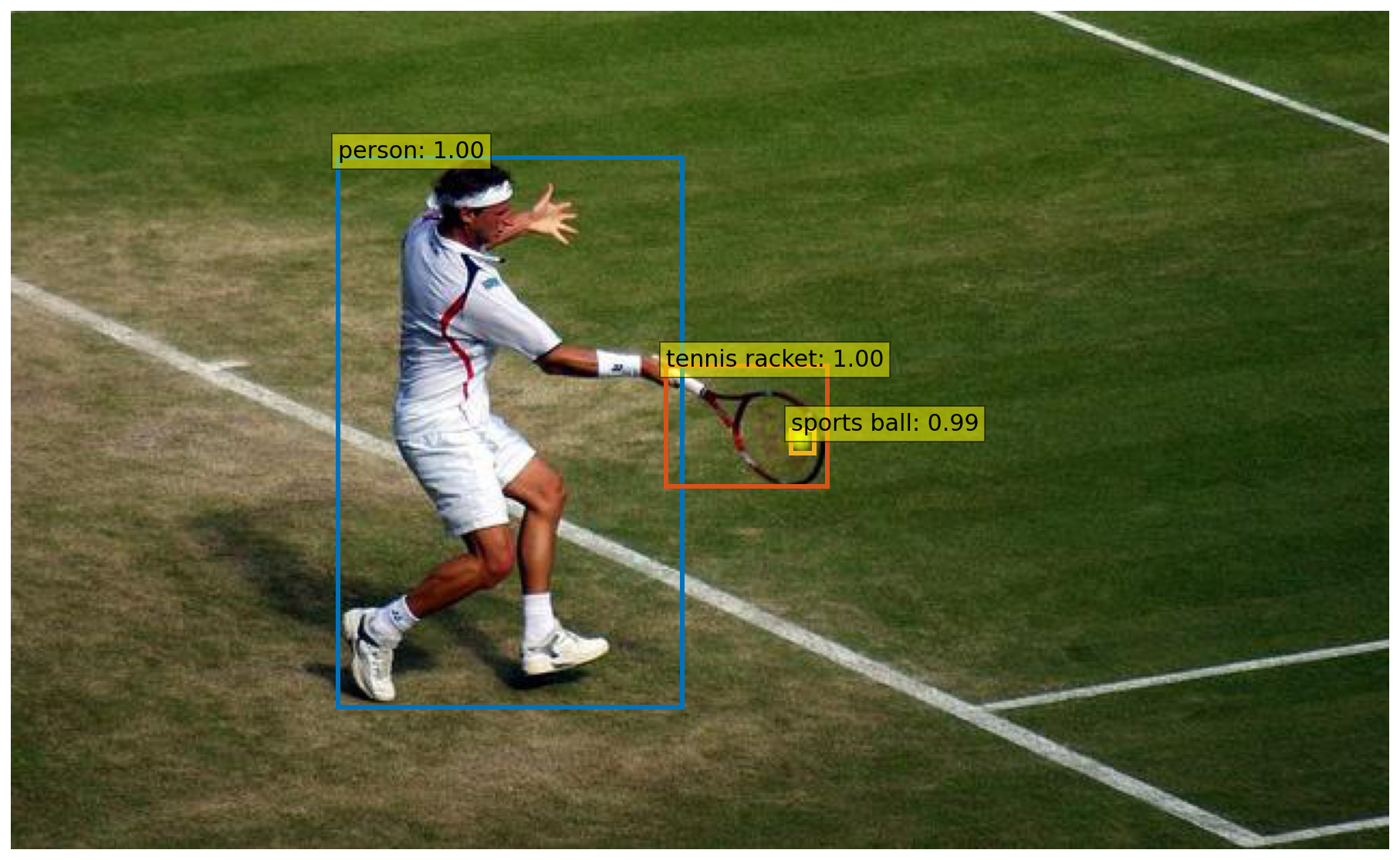}}%
\hfill
\raisebox{-.5\height}{\includegraphics[width=4.1cm]{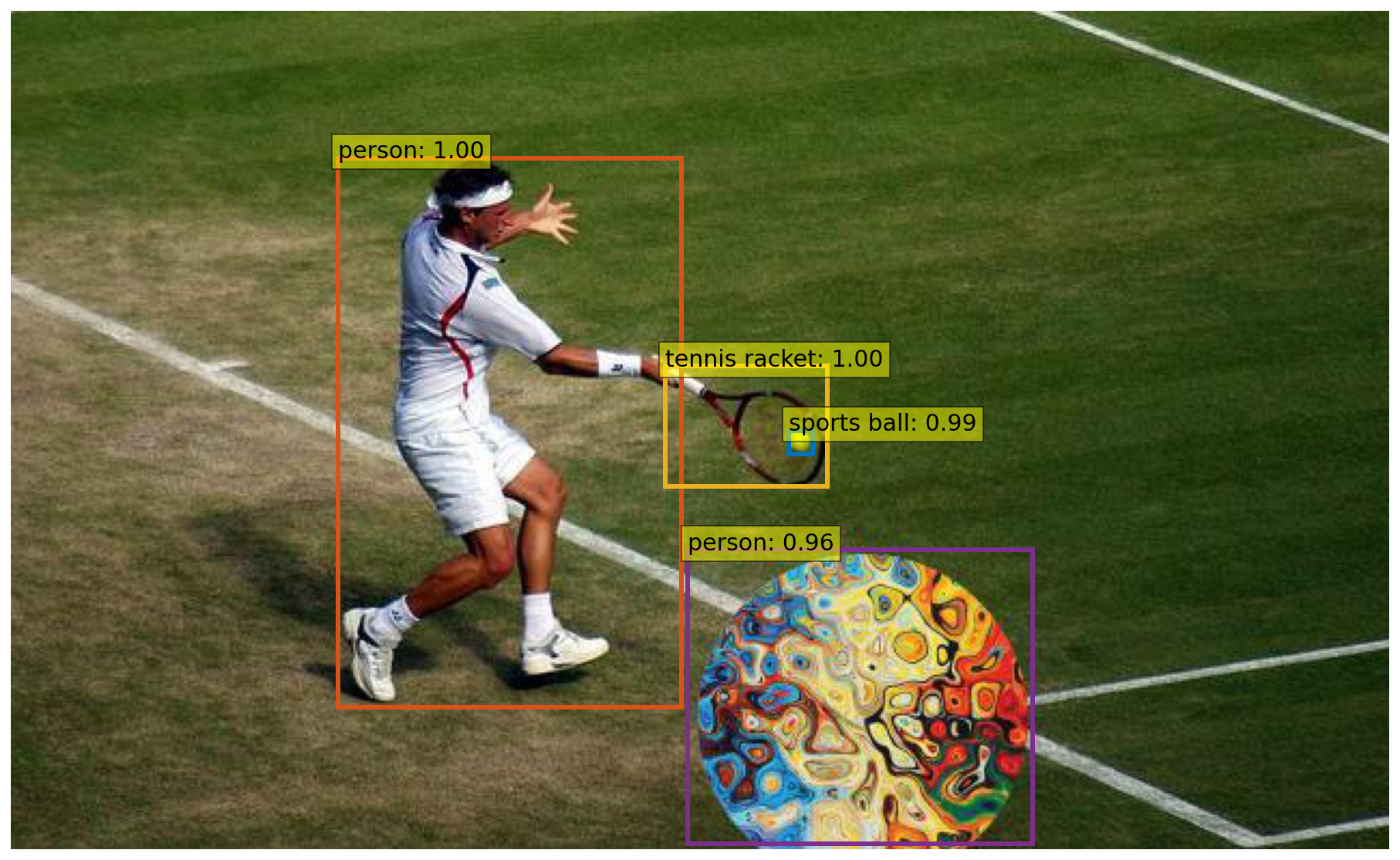}}%
\hfill
\raisebox{-.5\height}{\includegraphics[width=4.1cm]{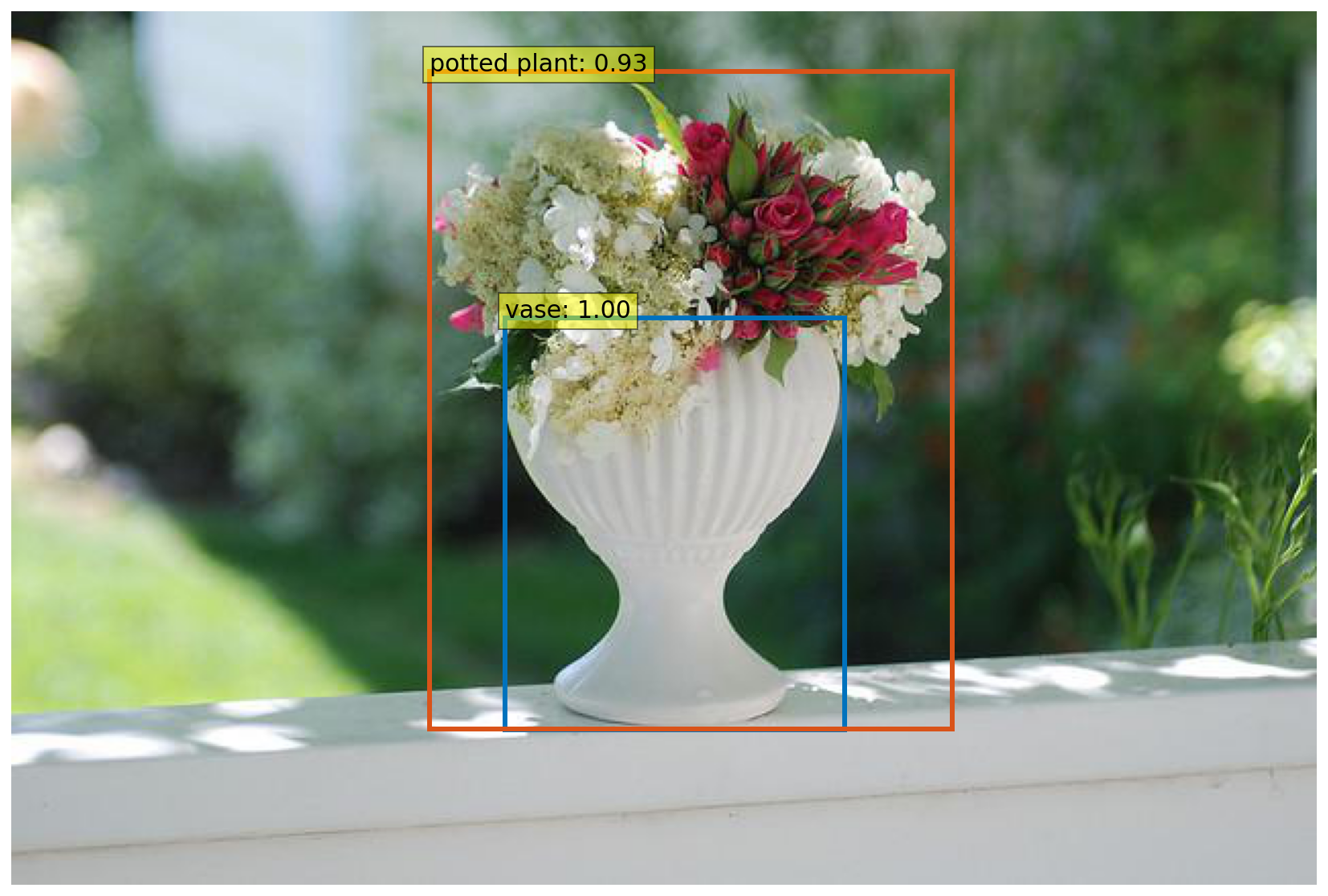}}%
\hfill
\raisebox{-.5\height}{\includegraphics[width=4.1cm]{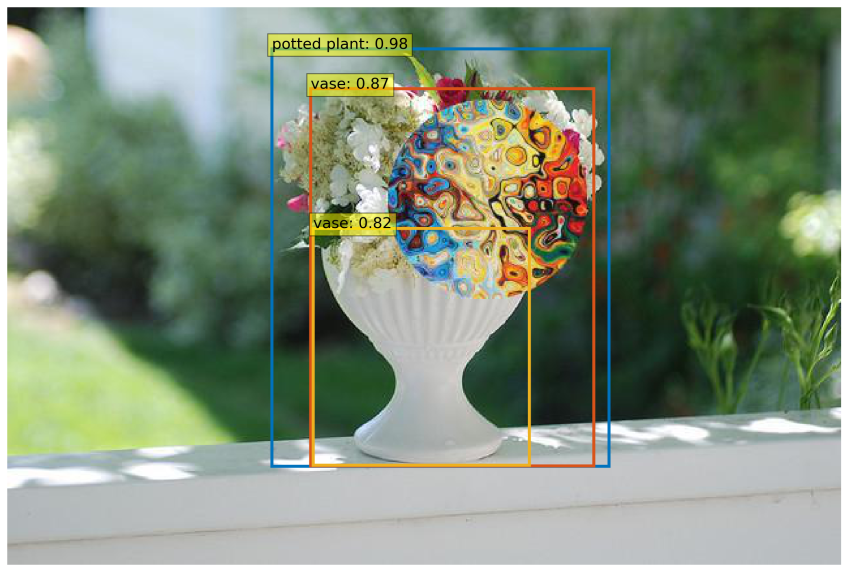}}%
\par
\caption{DETR detection result on original image (left) compare with the image with the adversarial sticker (right). The upper row shows the case where the sticker not overlapping other items in the image, while the lower row illustrated the opposite case.}
\label{fig:detr_sticker}
\end{figure}

We evaluate the three object detectors on images with adversarial stickers in which the detectors utilize ResNet101 as the backbone. For the dataset, we select 128 images from the MS COCO 2017 validation set \cite{d}. The adversarial patch was placed in the dataset images with random size and position. We report the resulting mAPs in Table \ref{tab:table_stickers}:  mAP under attacked images measured by the three object detectors. In terms of adversarial stickers evaluation, we find that Faster-RCNN gives the lowest performance among all three detectors, followed by DETR and YOLOv5.
\begin{table}[h!]
  \begin{center}
    \begin{tabular}{l|c|c}
      \textbf{} & 
      \textbf{mAP} & 
      \textbf{mAP$_{50}$ } \\
      \hline
      DETR & 0.512 & 0.726\\ 
      YOLOv5 & 0.548 & 0.743\\ 
      Faster-RCNN & 0.495 & 0.674\\
    \end{tabular}
    \caption{The Mean Average Precision (mAP) and mAP$_{50}$ result comparison of DETR, YOLOv5 and Faster-RCNN on images with adversarial stickers.}
    \label{tab:table_stickers}
  \end{center}
\end{table}
The resulting bounding box of DETR is also shown in Figure \ref{fig:detr_sticker}. As shown in the upper row of Figure \ref{fig:detr_sticker}, the network misclassified the attacked image, interpreting the stickers to be a person with 96\% confidence instead of entirely ignoring it. In the case the sticker not overlapping any item in the image, the sticker will not affect the accuracy of other detected objects in the image. However, in some cases where the stickers overlap the image, it can affect the attention process resulting in lower accuracy of other detected items in the image, as shown in the bottom row of Figure \ref{fig:detr_sticker}.

\begin{figure}[h]
    \centering
    \includegraphics[width=8.2cm]{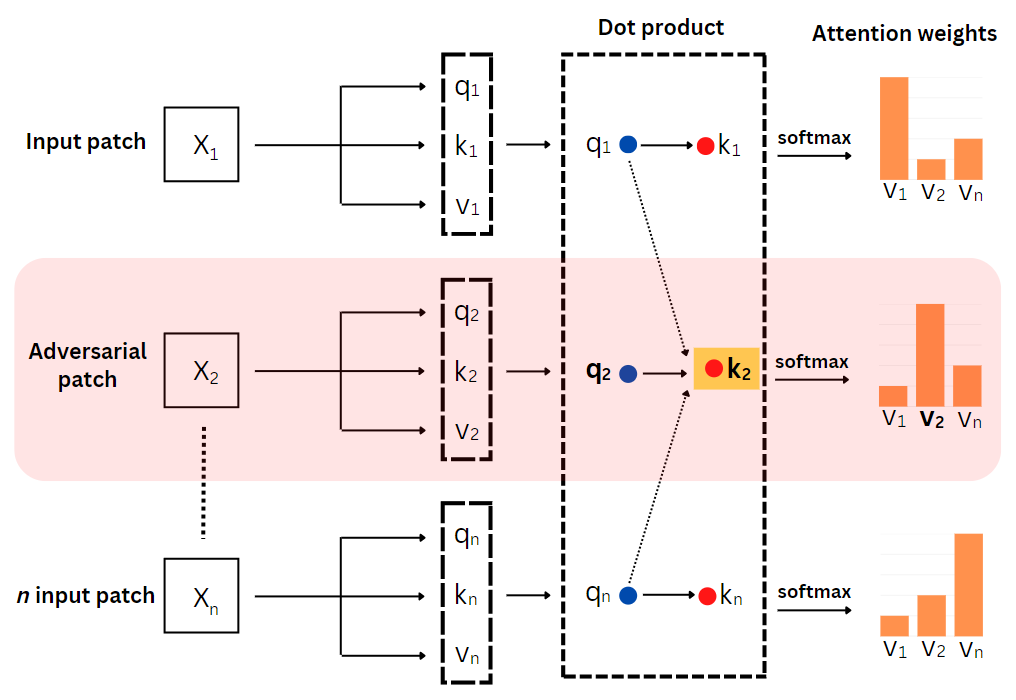}
    \caption{Self-attention mechanism for adversarial stickers patch settings. In this case, $q$, $k$, and $v$ represent the projected queries, keys, and value tokens of the input features. The mechanism involving dot-product attention computing the dot-product between queries with all corresponding keys before normalizing with softmax function to get the token attention weights. The adversarial patch introduces a new input feature at $X_2$ which misdirects the network's attention to the adversarial patch.}
    \label{fig:att_with_patch}
\end{figure}

This experiment implies that the sticker is capable of producing new unnecessary input features to be processed in the self-attention mechanism, which misdirects the network’s attention. The impact of the sticker on the self-attention mechanism of DETR is illustrated in Figure \ref{fig:att_with_patch}.
\hspace*{0.45cm}As can be seen in Figure \ref{fig:att_with_patch}, the sticker patch manages to produce a new set of keys, queries, and values in the network, which in most cases, results in the misdirection of the network. In the case where the stickers not overlapping other items in the image, the dot-product attention only computes the query $q_n$ with the corresponding key $k_n$. However, in most cases when the stickers overlap other items in the image, the queries of those particular items (i.e., $q_1, \dots, q_n$) are misdirected to the key token that represents the adversarial sticker (in this case the $k_2$). This result in increasing the attention weight to the corresponding adversarial patch key token. In other words, the network misguided the attention from actual image content to the adversarial stickers.
 \begin{figure}[h]
    \centering
    \includegraphics[width=8cm]{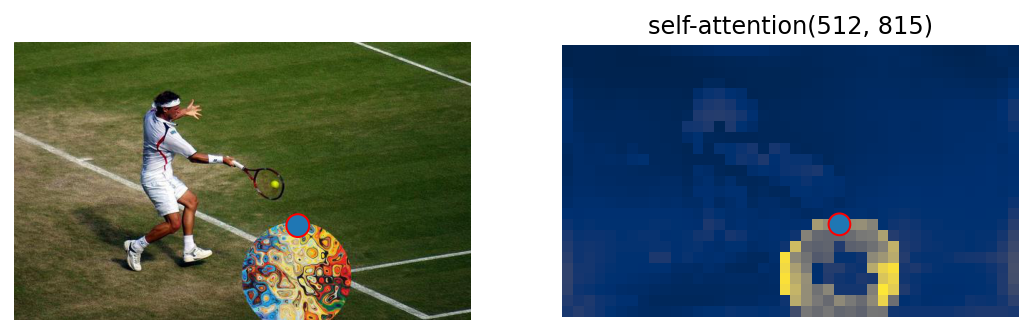}
    \caption{Visualization of self-attention weight on the adversarial patch.}
    \label{fig:visualization-attention}
\end{figure}
To gain intuition on what the detector see on a particular attacked image, we visualize the self-attention weight on the adversarial patch. As shown in the Figure \ref{fig:visualization-attention}, the network consider the entire sticker object to make prediction, counting it as the important object to predict.

\subsection{Evaluation on Image Corruption Benchmark}
\begin{figure}[h]
\par
\raisebox{-.5\height}{\includegraphics[width=8.3cm]{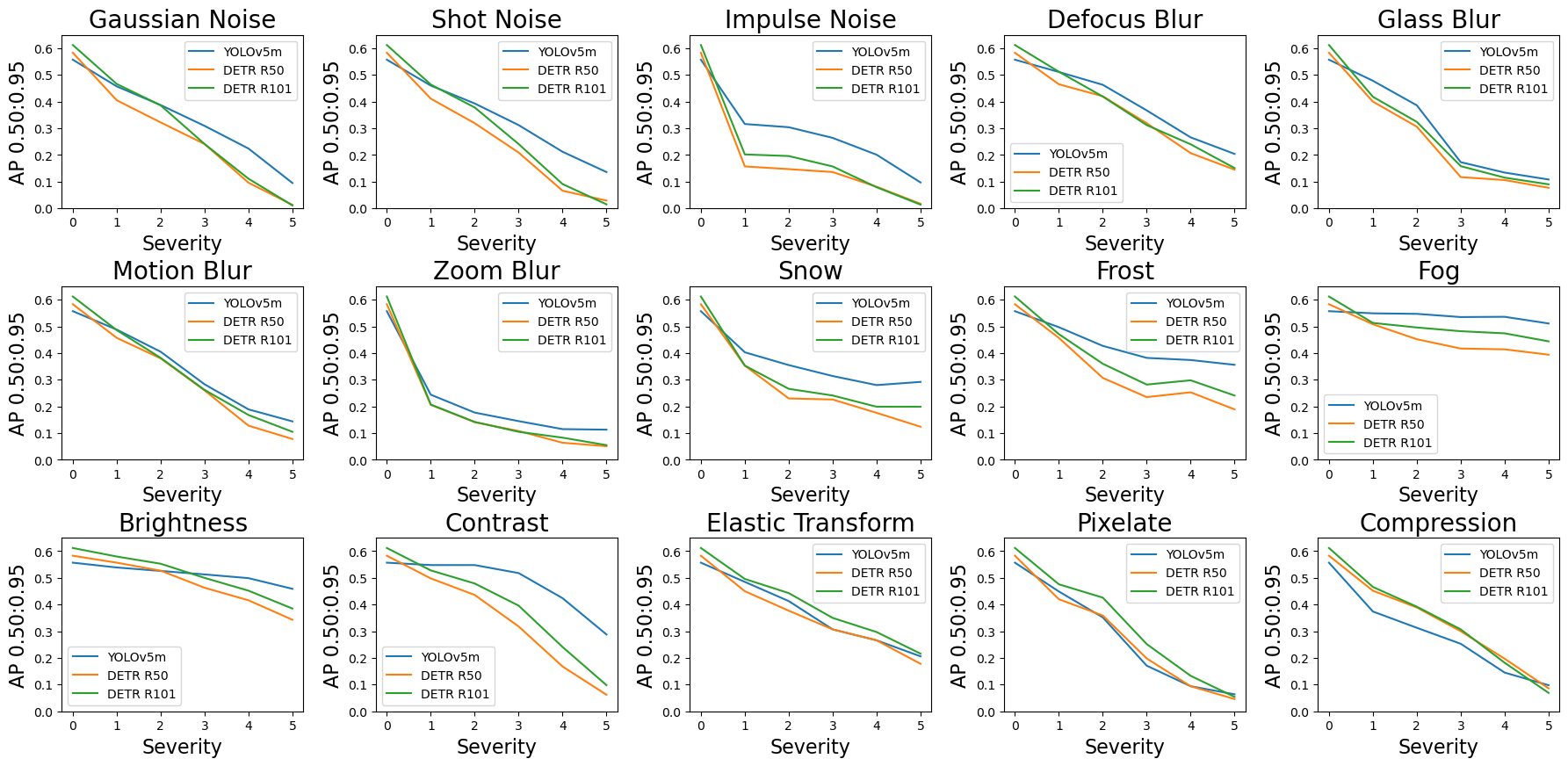}}%
\par
\caption{The benchmark results for three object detectors.}
\label{fig:bm_result}
\end{figure}

\begin{figure}[h]

\par
\raisebox{-.5\height}{\includegraphics[width=4.1cm]{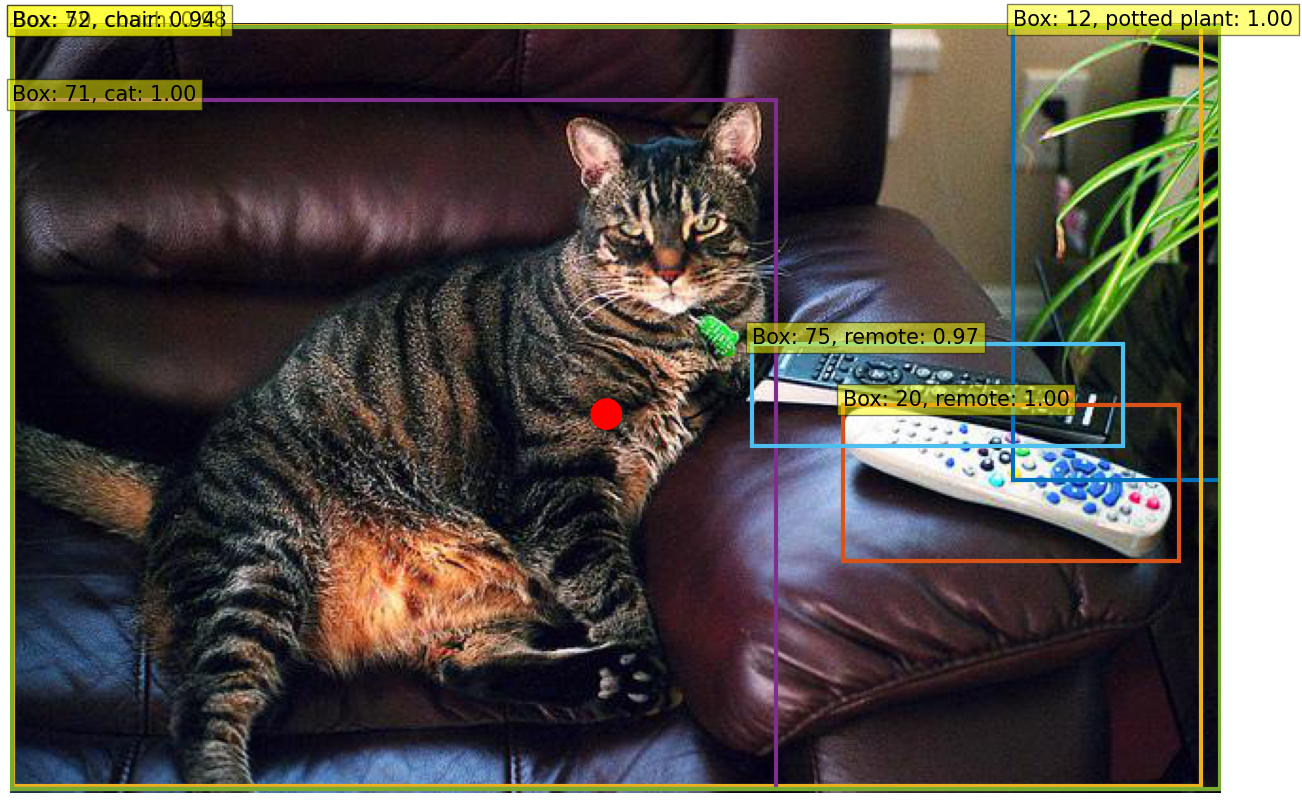}}%
\hfill
\raisebox{-.5\height}{\includegraphics[width=4.1cm]{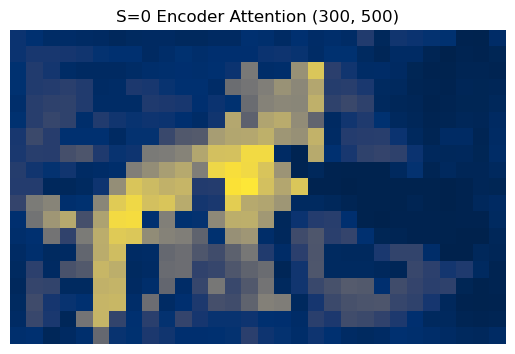}}%
\hfill
\raisebox{-.5\height}{\includegraphics[width=4.1cm]{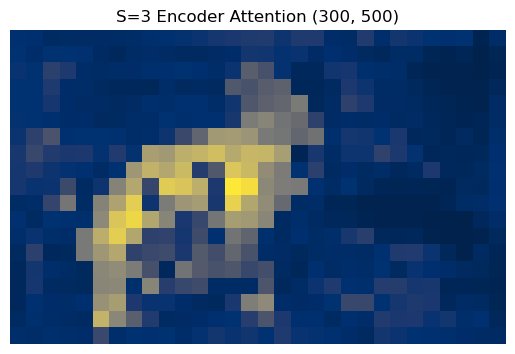}}%
\hfill
\raisebox{-.5\height}{\includegraphics[width=4.1cm]{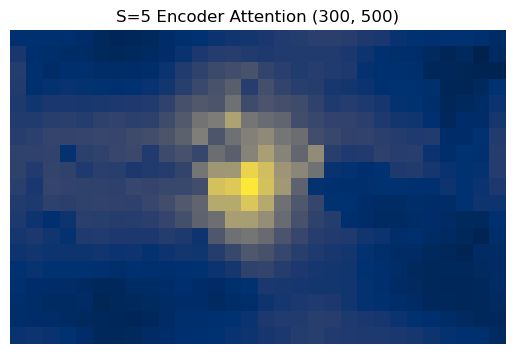}}%
\par
\caption{The detection result from DETR on original image and and encoder attention at (300, 450) as the impulse noise became more severe.}
\label{fig:bm_att}
\end{figure}

We evaluated the average precision of YOLOv5m, DETR R50, and DETR R101 on 15 image corruption categories and five levels of severity each, as shown in Figure \ref{fig:bm_result}. Although DETR models have better performance on original images, their precision on corrupted images is generally lower than YOLOv5m. DETR models performed much worse than YOLOv5m on images with impulse noise. Since DETR only used ReLU activation, it weakened the model's ability to counter these extreme values. We visualized the attention maps at pixel (300, 450) throughout the impulse noise test, shown in Figure \ref{fig:bm_att}. At the start, pixels corresponding to the "cat" had a strong correlation with our center point. However, as the impulse noise increases, the attention map gradually shrinks to the pixels near the center points. The encoder failed to correlate with "cat" pixels. This finally leads to a decrease in precision.

\subsection{Query Properties}
\begin{figure}[h]
\par
\raisebox{-.5\height}{\includegraphics[width=8.3cm]{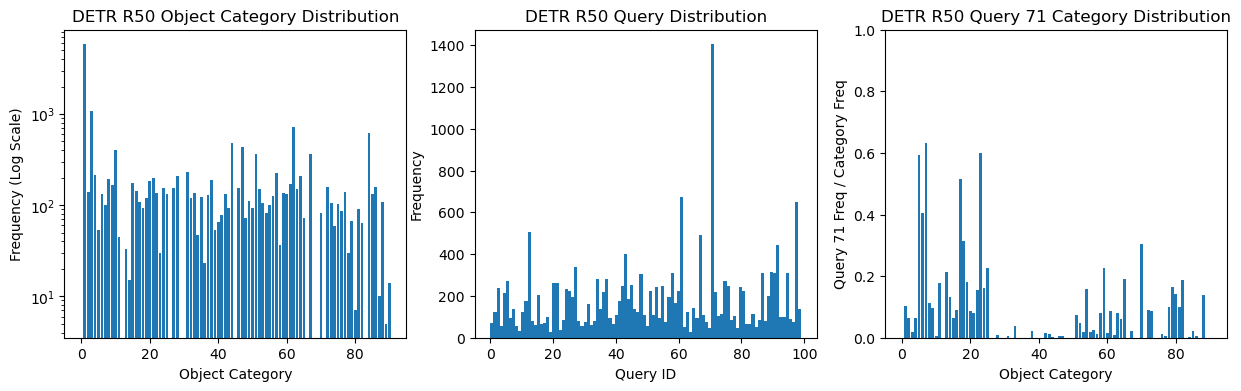}}%
\par
\raisebox{-.5\height}{\includegraphics[width=8.3cm]{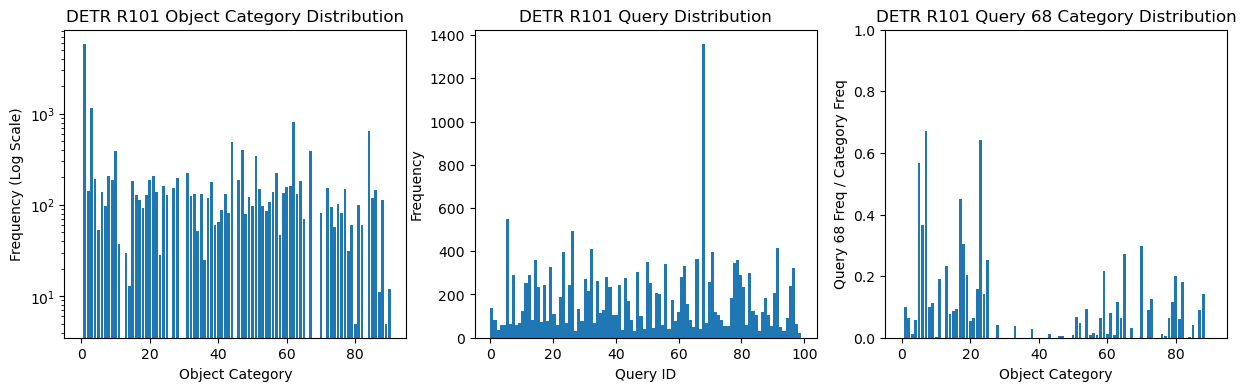}}%
\par
\caption{Left columns: Class distribution from model's prediction. Middle columns: Frequency of each query in predictions. Right columns: Main queries' contribution in detecting each object category.}
\label{fig:query_freq}
\end{figure}

First, we evaluated the contribution of each query in object detection. Both DETR R50 and DETR R101 models were tested on the MSCOCO validation dataset. We collected all predictions with confidence larger than 0.8 and the corresponding query ID to compute the query frequency, as shown in the middle plots in Figure \ref{fig:query_freq}. We found an interesting phenomenon that both the DETR R50 and DETR R101 model have one main query (71 in DETR R50 and 68 in DETR R101) that detect lots of objects. Each main query can account for 7.5\% of the total predictions. Intuitive thinking was that the main queries are responsible for detecting "person", which is the most common object in the dataset. Therefore, we investigated the relations between main queries and classes. We divided the main queries' frequency by the total frequency in each class to visualize their contribution, as shown in the right plots. Both main queries have shown very similar distribution over object categories. None of them has a high contribution ($<$ 40\%) in the "person" category, and both of them has a high contribution ($<$ 40\%) in the "airplane", "train", "cat" and "bear" category. This indicated that the main query has particular preferences on object classes, but they are not detecting the most common object, "person".

\begin{figure}[h]
\par
\raisebox{-.5\height}{\includegraphics[height=4cm, width=8.3cm]{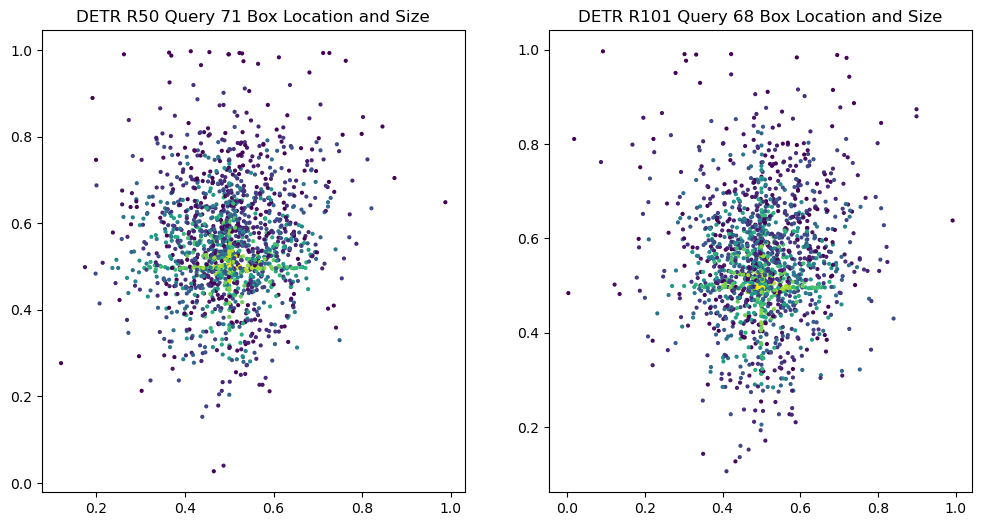}}%
\par
\caption{The predicted bounding boxes center location and size of main queries. Yellow colored points refers to boxes with a large size.}
\label{fig:query_loc_size}
\end{figure}

Apart from class relations, we also analyzed bounding boxes' location and size predictions from main queries, shown in Figure \ref{fig:query_loc_size}. Both queries prefer to detect medium to large objects in the center region. However, there are also a large amount of small to medium-sized boxes circulating the center.

To analyze the importance of the main query, we masked its outputs and evaluated the precision change, as shown in Table 3. The main query indeed has a massive influence on the model's performance. Without predictions from the main query, the average precision dropped by around 7 points. This shows that DETR is heavy depending on the predictions from the main query. It will make the model more vulnerable to attacks.

\begin{table}[h!]
  \begin{center}
    \label{tab:query}
    \begin{tabular}{l|c|c}
      \textbf{} & 
      \textbf{mAP} & 
      \textbf{mAP without main query} \\
      \hline
      DETR R50 & 0.420 & 0.353\\ 
      DETR R101 & 0.435 & 0.367\\ 
    \end{tabular}
    \caption{The mAP of two models with and without predictions from the main query.}
  \end{center}
\end{table}

\subsubsection{Cause of The Main Query}
\begin{figure}[h]
\par
\raisebox{-.5\height}{\includegraphics[width=4.1cm]{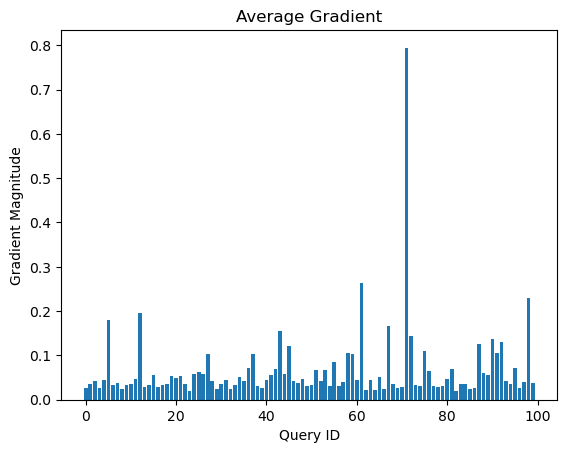}}%
\hfill
\raisebox{-.5\height}{\includegraphics[width=4.1cm]{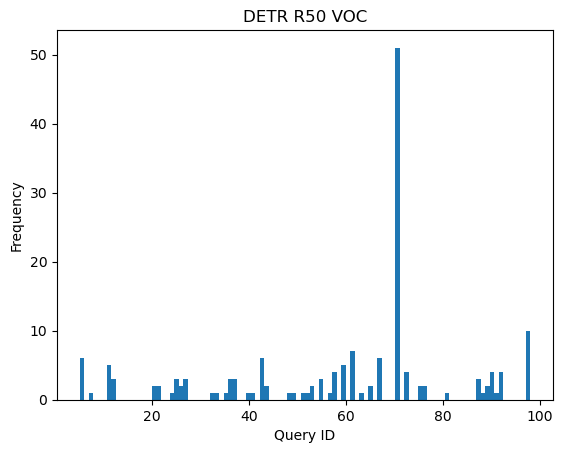}}%
\par
\caption{The average gradient flow to each query during transfer learning and the query frequency in pascal-voc-dataset.}
\label{fig:query_grad}
\end{figure}

In the following experiment, we investigated the cause of the main query and the imbalanced query contribution problem in the DETR model. We fine-tuned a DETR R50 model on the pascal-voc dataset. We first trained the model for 50 epochs to let it stabilize and then collected the gradient information that was applied to each query in the following ten epochs. In Figure \ref{fig:query_grad}, we computed the mean gradient magnitude applied on each query and the query frequency. Query 71 remained as the main query in pascal-voc dataset. It received the most gradient flow, and its magnitude is significantly larger than any other query. We believe the cause for the imbalanced gradient flow is due to the transformer structure. The DETR wanted to keep its transformer module simple so every query is allowed to compute cross-attentions with the whole encoder context. In the early stage of training, the main query converged fast and made the best predictions among all queries though it might not be precise. However, this is enough for the main query to receive most of the gradient flow, while others had much fewer chances to update and compete with the main query. It not only forced this single query to converge to most of the detection tasks, but also affected other queries' convergence progress. Therefore, the imbalanced query contribution problem caused the slow convergence and low performance of the DETR model.

\subsubsection{Solutions to The Main Query}
Our ideas against this problem are simple and direct, to suppress positive predictions from the main query and encourage that from all remaining queries. This can be achieved by two methods: random query drop and cross-attentions through a smaller window.

\begin{figure}[h]
\par
\raisebox{-.5\height}{\includegraphics[height=6cm, width=8.2cm]{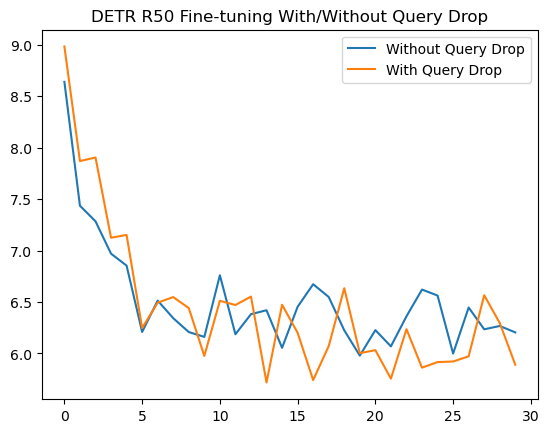}}%
\par
\caption{The loss graph comparison between two fine-tuned model with and without random query drop.}
\label{fig:query_loss}
\end{figure}

To evaluate the performance of random query drop, we fine-tuned another DETR R50 model on pascal-voc dataset for 30 epochs and compared the loss graph with the previous fine-tuned model, as shown in Figure \ref{fig:query_loss}. We observed that the model with random query drop has a lower testing loss and showed faster convergence. Random query drop is only a prototype idea proposed by us, and we hope it can be further evaluated on other transformers that use independent queries like DETR. The second idea is similar to the one proposed in Swin Transformer \cite{qu_1}, where attention is computed inside a small window. This not only reduces the computation cost but also prevents any single query from being able to answer all detection tasks.

\section{Conclusion}
In this work, we investigate the robustness and properties of transformer-based detectors DETR by measuring its performance toward different image nuisances and making a comparison with other detectors. Two widely adopted CNN-based detectors like YOLOv5 and Faster-RCNN, are used to compare their performance with DETR. These detectors are evaluated with two main test cases, occlusion and adversarial stickers. We implement random and salient occlusion to analyze how the network handles the specific regions impacted by occlusion, specifically on the region where it contains essential information in the image. In terms of adversarial stickers, this attack is utilized to change the pixel values in some portion of the image to mislead the network. Moreover, we benchmark the robustness of DETR and analyze the query properties to a set of corruption images.

Our experiment implies that DETR performs well when it comes to resistance to interference from information loss in occlusion images. Nevertheless, in the case of the sticker patch, it manages to produce a new set of keys, queries, and values in the network which in most cases results in the misdirection of the network. This result in increasing the attention weight to the corresponding adversarial patch key token, making the network misguided the attention to the adversarial stickers. For the benchmark, the experiment indicates that DETR precision on corrupted images is generally lower than YOLOv5. We also found that DETR depends heavily on the predictions from the main query. This impact the main query to receive most of the gradient flow, which leads to imbalanced contributions among all queries. 

{\small


}
\end{document}